\documentclass[letterpaper, 10 pt, conference]{ieeeconf}  % Comment this line out if you need a4paper

\IEEEoverridecommandlockouts                      
\usepackage{xcolor, soul}
\usepackage{subcaption}
\usepackage{amsmath}
\usepackage{amsfonts}
\usepackage{amssymb}
\usepackage{bm}
\usepackage{multicol}
\usepackage{multirow}
\usepackage{optidef}
\usepackage{scalerel}
\usepackage{tikz}
\usetikzlibrary{svg.path}
\usepackage{colortbl}
\usepackage{color}
\definecolor{celadon}{rgb}{0.67, 0.88, 0.69}
\usepackage{algorithm}
\usepackage{algpseudocode}
\usepackage{lipsum}
\usepackage{titlesec}
\usepackage{hyperref}

\definecolor{orcidlogocol}{HTML}{A6CE39}
\tikzset{
  orcidlogo/.pic={
    \fill[orcidlogocol] svg{M256,128c0,70.7-57.3,128-128,128C57.3,256,0,198.7,0,128C0,57.3,57.3,0,128,0C198.7,0,256,57.3,256,128z};
    \fill[white] svg{M86.3,186.2H70.9V79.1h15.4v48.4V186.2z}
                 svg{M108.9,79.1h41.6c39.6,0,57,28.3,57,53.6c0,27.5-21.5,53.6-56.8,53.6h-41.8V79.1z M124.3,172.4h24.5c34.9,0,42.9-26.5,42.9-39.7c0-21.5-13.7-39.7-43.7-39.7h-23.7V172.4z}
                 svg{M88.7,56.8c0,5.5-4.5,10.1-10.1,10.1c-5.6,0-10.1-4.6-10.1-10.1c0-5.6,4.5-10.1,10.1-10.1C84.2,46.7,88.7,51.3,88.7,56.8z};
  }
}

\newcommand\orcidiconWF[1]{\href{https://orcid.org/XXXX-XXXX-XXXX-XXXX}{\mbox{\scalerel*{
\begin{tikzpicture}[yscale=-1,transform shape]
\pic{orcidlogo};
\end{tikzpicture}
}{|}}}}

\newcommand\orcidiconKLW[1]{\href{https://orcid.org/0000-0002-1938-4222}{\mbox{\scalerel*{
\begin{tikzpicture}[yscale=-1,transform shape]
\pic{orcidlogo};
\end{tikzpicture}
}{|}}}}

\newcommand\orcidiconYY[1]{\href{https://orcid.org/0000-0002-5797-9753}{\mbox{\scalerel*{
\begin{tikzpicture}[yscale=-1,transform shape]
\pic{orcidlogo};
\end{tikzpicture}
}{|}}}}

\newcommand\orcidiconFGS[1]{\href{https://orcid.org/0000-0002-5090-9007}{\mbox{\scalerel*{
\begin{tikzpicture}[yscale=-1,transform shape]
\pic{orcidlogo};
\end{tikzpicture}
}{|}}}}

 \title{\LARGE \bf
 Computing forward statics from tendon-length in flexible-joint hyper-redundant manipulators}

\author{Weiting Feng$^{1}$, Kyle L. Walker$^{2}$\orcidiconKLW{}, Yunjie Yang$^{3}$\orcidiconYY{}, and Francesco Giorgio-Serchi$^{1}$\orcidiconFGS{}% <-this % stops a space
\thanks{$^{1}$Weiting Feng and Francesco Giorgio-Serchi are with the Institute for Integrated Micro and Nano Systems, School of Engineering, University of Edinburgh, Edinburgh,
U.K. (Correspondence: F.Giorgio-Serchi@ed.ac.uk).}% <-this % stops a space
\thanks{$^{2}$Kyle L. Walker is with the CREATE Lab, EPFL, Lausanne, Switzerland.}% <-this % stops a space
\thanks{$^{3}$Yunjie Yang is with the Institute for Imaging, Data and Communications, School of Engineering, University of Edinburgh, Edinburgh, UK.}%
}

\begin{document}

\maketitle
\thispagestyle{empty}
\pagestyle{empty}

%%%%%%%%%%%%%%%%%%%%%%%%%%%%%%%%%%%%%%%%%%%%%%%%%%%%%%%%%%%%%%%%%%%%%%%%%%%%%%%%
\begin{abstract}

Hyper-redundant tendon-driven manipulators offer greater flexibility and compliance over traditional manipulators. A common way of controlling such manipulators relies on adjusting tendon lengths, which is an accessible control parameter. This approach works well when the kinematic configuration is representative of the real operational conditions. However, when dealing with manipulators of larger size subject to gravity, it becomes necessary to solve a static force problem, using tendon force as the input and employing a mapping from the configuration space to retrieve tendon length. Alternatively, measurements of the manipulator posture can be used to iteratively adjust tendon lengths to achieve a desired posture. Hence, either tension measurement or state estimation of the manipulator are required, both of which are not always accurately available. Here, we propose a solution by reconciling cables tension and length as the input for the solution of the system forward statics. We develop a screw-based formulation for a tendon-driven, multi-segment, hyper-redundant manipulator with elastic joints and introduce a forward statics iterative solution method that equivalently makes use of either tendon length or tension as the input. This strategy is experimentally validated using a traditional tension input first, subsequently showing the efficacy of the method when exclusively tendon lengths are used. The results confirm the possibility to perform open-loop control in static conditions using a kinematic input only, thus bypassing some of the practical problems with tension measurement and state estimation of hyper-redundant systems.

\end{abstract}

%%%%%%%%%%%%%%%%%%%%%%%%%%%%%%%%%%%%%%%%%%%%%%%%%%%%%%%%%%%%%%%%%%%%%%%%%%%%%%%%
\section{Introduction}

Hyper-redundant and flexible manipulators, often characterized by their high degrees of freedom, have been proposed in a multitude of industries for offering greater compliance, dexterity and a greatly expanded workspace in comparison to rigid-link counterparts \cite{Mu2022, Russo2023}. However, this increased adaptability introduces significant challenges in accurately determining the configuration of such manipulators \cite{478427}. Unlike traditional rigid-link designs that only feature a low number of joints, where joint positions and end-effector locations can be determined with relatively straightforward kinematic models, flexible and hyper-redundant manipulators exhibit highly complex and nonlinear behaviors \cite{Chirikjian1994}. Furthermore, tendon-driven actuation is a common practice, mainly due to the fact that tendon-length is a very accessible control parameter to observe. However, tendon-driven actuation also has some drawbacks, such as increased kinematic complexity \cite{Murray1994} and the lack of simple methods for performing open-loop control under static force conditions. 
\begin{figure}[t!]
    \centering
    \includegraphics[width = 0.95\columnwidth ]{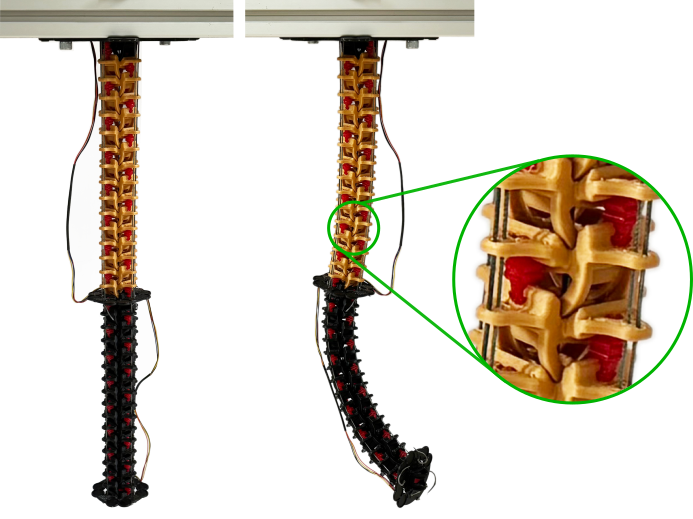}
    \caption{A hyper-redundant, flexible-joint, tendon-driven manipulator with two independent segments consisting of a sequence of rigid beads, connected by elastic hinges \cite{Walker2025}. Two successive joints can rotate perpendicularly to each other; see magnified inset.).}
    \label{fig:side-by-side}
    \vspace{-0.7cm}
\end{figure}
Reliance on kinematic models remains a widespread control approach that offers analytical solutions to estimate posture based on predefined shape assumptions \cite{Jones2006, Webster2010, Allen2020}. However, adopting a purely kinematic approach fails to consider force information, which ultimately controls the posture of the robot. This becomes problematic for large-scale manipulators subject to gravity and other restoring forces, as the kinematic solution diverges from a static one. 

Circumventing the limits of kinematic models requires solving of a static problem, for which tendon forces are required. There are well-established methods for computing the forward statics from tendon-force inputs \cite{Rucker2011, Rao2021}, assuming a continuous and uniform backbone stiffness. However, in hyper-redundant systems such as the one of Fig. \ref{fig:side-by-side}, the segment-wise variations in stiffness and curvature can easily diverge from the continuum assumption. Even for approximations that don't rely on the constant strain assumption \cite{Renda2020}, practical implementation still requires a method to measure tendon force. Regardless of the model used, tendon forces are not always easily accessible experimentally, unlike tendon lengths. Also, retrieving tendon length from a static model requires a mapping from configuration space to actuation space and for this, accurate information of the spatial configuration of the whole manipulator is necessary. Hence, performing forward statics based on tendon lengths remains unsolved outside of prescribed assumptions \cite{Rone2013,Dalvand2018,Oliver2019}. Alternatively, closed-loop control allows bypassing of the static solution by iteratively converging towards a desired configuration, but once again it relies on accurate state estimation and mapping from configuration to actuation space.
%Common approaches to address these challenges have relied on the use of inverse iterative models or iterative numerical techniques to estimate configuration based on tendon forces, often employing simplifications such as a continuum approximation \cite{Chirikjian1994b}. 
While effective, these approaches require accurate state estimation which is an open problem in soft and hyper-redundant systems, often being imprecise in practice or altogether not accessible \cite{doi:10.1126/scirobotics.aav1488}.

\begin{figure*}[t!]
    \centering
    \includegraphics[width = 0.98\textwidth ]{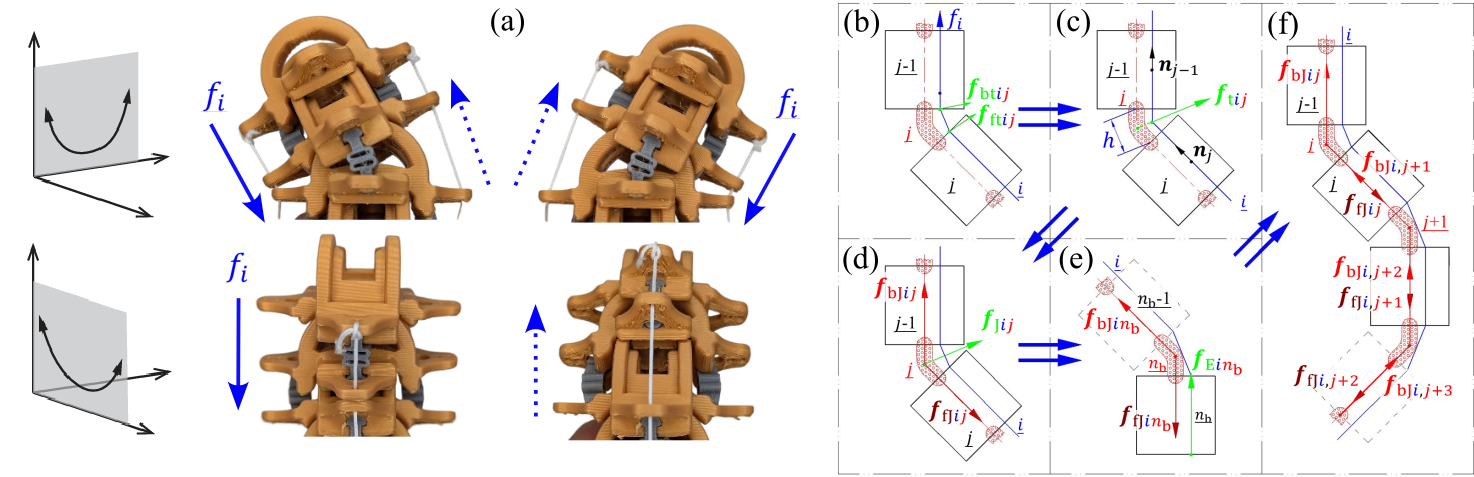}
    \caption{(a) Orthogonal motion of successive beads when a force $f_i$ is applied to a tendon and (b)-(f) a schematic diagram of the Elastic-joint model, depicting beads as a square connected by a flexible hinge (dotted red area). %A hyper-redundant, flexible-joint, tendon-driven manipulator with two independent segments. Each segment is actuated by four partially-constrained tendons. This manipulator consists of a sequence of rigid beads, identified by the yellow elements, (b) connected by elastic hinges, identified by the red units, which define a joint. Two successive joints can rotate perpendicularly to each other, as shown in (c).
    }
    \label{fig:elastic}
    \vspace{-0.7cm}
\end{figure*}

%Also, for tendon-driven actuation, joint compliance, friction and hysteresis can lead to discrepancies between expected and actual configurations, even with a closed-loop feedback loop \cite{Liu2021}.
With respect to this, this paper presents a computational framework for evaluating the open-loop forward statics problem of tendon-driven flexible-joint hyper-redundant manipulators, by taking cables length, rather than cables tension, as the input. The proposed approach specifically addresses cases where stiffness is distributed across multiple joints per segment, using a screw-based formalism to efficiently treat the multitude of joints. This approach lends itself to the modeling of a broad family of both tendon-driven continuum \cite{Rao2021} and discrete manipulators, treating the system as a discrete sequence of multiple stacked compliant joints analogous to that in \cite{Kato2015}. The proposed solution strategy of equivalently making use of tendon-lengths and tendon tensions as an input to compute the resulting static configuration of the manipulator, accounting for gravity, distributed stiffness and other internal forces such as discontinuous tendon routing. Also, continuity with previous methods is maintained by being able to extract tendon lengths or tensions as an output when the model is provided with only one of these input parameters. 
%The method is designed to use a kinematic input to solve a forward statics problem via screw theory, aiming to improve computational efficiency while maintaining accuracy. 
The model is experimentally validated to demonstrate the effectiveness of the approach, utilizing a representative manipulator platform featuring two independently actuated segments constructed of an array of flexible joints.

This paper is structured as follows: first, a screw-based formulation for the static of a tendon-driven, flexible joint manipulator is introduced, highlighting the coupled solution approach with tendon force and tendon length as the input to the forward static problem; second, the experimental platform used for validation is described, which offers accurate tension force measurement and optical state-estimation, making it suitable for validation of the model. First, the forward static solution based on tendon force input is validated, assessing the model performance against a kinematic piecewise constant curvature approximation. Then, we show statics solution achieved by taking tendon length as the only input and computing tendon force as a byproduct. The results confirm the framework’s ability to capture the statics solution of tendon-driven hyper-redundant manipulators without relying on cable tension measurements. 

%this is an example of how to include a figure
%\begin{figure}[t!]
    %\centering
    %\includegraphics[width=0.48\textwidth]{figures/traj_art_schematic.png}
    %\caption{Example of a figure.}
    %\label{art_schematic}
%\end{figure}

%%%%%%%%%%%%%%%%%%%%%%%%%%%%%%%%%%%%%%%%%%%%%%%%%%%%%%%%%%%%%%%%%%%%%%%%%%%%%%%%

\section{Modelling}
%\subsection{Basic definitions from screw theory}
Following rigid body kinematics, we use the homogeneous transformation matrix $\bm T$ to represent the posture of a reference frame:
\begin{equation}
\boldsymbol{T} =
\begin{bmatrix}
\boldsymbol{R} & \boldsymbol{p} \\
\boldsymbol{0} & 1
\end{bmatrix}
\end{equation}
with \( \boldsymbol{R} \in SO(3) \) the rotation matrix, and \( \boldsymbol{p} \in \mathbb{R}^3\)  the position.

%The set of postures of a rigid body is the special Euclidean group \( SE(3) \). As a Lie group, \( SE(3) \) can naturally find its Lie algebra \( \mathfrak{se}(3) \). 
We define the twist $\mathcal{V}$ and $\mathcal{S}$ the screw axis such that:
\begin{equation}\label{screw_1}
{}^{\ast}\boldsymbol{\mathcal{V}} =
\begin{bmatrix}
{}^{\ast}\boldsymbol{\omega} & \boldsymbol{v} \\
\boldsymbol{0} & 0
\end{bmatrix}, 
{}^{\ast}\boldsymbol{\mathcal{S}} = 
\dfrac{{}^{\ast}\boldsymbol{\mathcal{V}}}{||\boldsymbol{\omega}||}
\end{equation}
where \( {}^{\ast}\boldsymbol{\omega} \) is the anti-dual form (see appendix \ref{appendix_A}) of the angular velocity vector \( \boldsymbol{\omega} \), and \( \boldsymbol{v} \) is the linear velocity vector in. By default these two vectors are defined in the body frame.
% \( ||\boldsymbol{\omega}|| \) is the norm of \( \boldsymbol{\omega} \).

% \begin{figure*}[t!]
%     \centering
%     \includegraphics[width =\columnwidth ]{figures/elastic_joint_zoom.png}
%     \caption{Schematics of the manipulator joint with force breakdown: (a) the case where the tendon pass externally and (b) the case where the tendons pass internally. \hl{(Note: here we need an additional image to show the real joint and where the cables are}}
%     \label{fig:side-by-side}
% \end{figure*}
Based on these definition, the following exponential mapping holds:
\begin{equation}
\boldsymbol{T} = \text{exp}({}^{\ast}\boldsymbol{\mathcal{V}}t) = \text{exp}({}^{\ast}\boldsymbol{\mathcal{S}}\theta)
\end{equation}
where \(t\) is the duration of the motion, and \( \theta \) is the angle covered by the motion. From eq. \ref{screw_1}, twist and screw axis can be expressed in vector form as,
\begin{equation}
\boldsymbol{\mathcal{V}} = 
\begin{bmatrix}
\boldsymbol{\omega}\\
\boldsymbol{v} 
\end{bmatrix} \in \mathbb{R}^6,
\boldsymbol{\mathcal{S}} = 
\frac{\boldsymbol{\mathcal{V}}}{||\boldsymbol{\omega}||} \in \mathbb{R}^6,
\end{equation}
%Because the magnitude of angular velocity is removed from screw, it can represent the screw axis itself, while twist represents the specific spiral motion around that axis.
%Similarly, we can combine force and moment into a vector, called a force screw:
from which follows that a wrench $\bm{\mathcal{F}}$ combining both linear force $\bm f$ and moment $\bm m$ is defines as,
\begin{equation}
\boldsymbol{\mathcal{F}} =
\begin{bmatrix}
\boldsymbol{m} \\
\boldsymbol{f}
\end{bmatrix} \in \mathbb{R}^6
\end{equation}
%where \( \boldsymbol{m} \) represents the moment and \( \boldsymbol{f} \) represents the force applied to the rigid body. According to Newton and Euler's laws of motion, we can get the relationship between the force screw acting on a rigid body and the twist of the rigid body:
Following \cite{Lynch}, the relationship between wrench and twist of a rigid body can be written as,
\begin{equation}
\boldsymbol{\mathcal{F}} = \boldsymbol{\mathcal{G}}\dot{\boldsymbol{\mathcal{V}}}-\text{ad}_{\boldsymbol{\mathcal{V}}} \boldsymbol{\mathcal{G}}\boldsymbol{\mathcal{V}}
\end{equation}
where \(\boldsymbol{\mathcal{G}}\) is the spatial inertia matrix consisting of the moment of inertia \(\boldsymbol{\mathcal{I}}\) and mass \(m\), and \(\text{ad}_{\boldsymbol{\mathcal{V}}}\) is the adjoint operator  of  \(\boldsymbol{\mathcal{V}}\) (see appendix \ref{appendix_B} for a formal definition).
\begin{equation}
\boldsymbol{\mathcal{G}}\ =
\begin{bmatrix}
\boldsymbol{\mathcal{I}} & \boldsymbol{0} \\
\boldsymbol{0} & m\boldsymbol{I}
\end{bmatrix}
\end{equation}
%\subsection{Dynamic of manipulators}
%A manipulator usually consists of multiple rigid bodies. Using eq6, we can get the close form of manipulator dynamics:
This allows us to write the general form of a manipulator's dynamics in the form:
\begin{equation}\label{dynamics}
\boldsymbol\tau =
\boldsymbol{M}(\theta)\ddot{\boldsymbol\theta}+\boldsymbol{c}(\boldsymbol\theta,\dot{\boldsymbol\theta})+\boldsymbol{g}(\boldsymbol\theta)+\boldsymbol{J}^T(\boldsymbol\theta)\boldsymbol{\mathcal{F}}_{\text{e}}
\end{equation}
where \( \boldsymbol\theta \) is joint vector, and \( \boldsymbol\tau \) is the active torque on all joints. \( \boldsymbol{M}(\boldsymbol\theta) \) is the mass matrix, \( \boldsymbol{c}(\boldsymbol\theta,\dot{\boldsymbol\theta}) \) is the combination of Coriolis and centripetal torque and \( \boldsymbol{g}(\boldsymbol{\theta}) \) is the gravitational torque. Here $\bm{\mathcal{F}}_e$ represent an external wrench at the end-effector and $\bm{J}(\bm{\theta})$ is the wrench Jacobian. By defining terms $\bm{\mathcal{A}}$, $\bm{\mathcal{W}}$ and $\bm{\mathcal{L}}$ according to appendix \ref{appendix_D}, these matrices can be succinctly expressed as follows,
\begin{equation}
\begin{cases}
\hfill{\boldsymbol{M}(\boldsymbol\theta)} &= \boldsymbol{\mathcal{A}}^T\boldsymbol{\mathcal{L}}^T\boldsymbol{\mathcal{G}}\boldsymbol{\mathcal{L}}\boldsymbol{\mathcal{A}} \\
\hfill{\boldsymbol{c}(\boldsymbol\theta,\dot{\boldsymbol\theta})} &= \boldsymbol{\mathcal{A}}^T\boldsymbol{\mathcal{L}}^T(\boldsymbol{\mathcal{G}}\mathcal{L}[\text{ad}_{\boldsymbol{\mathcal{A}}\dot{\boldsymbol\theta}}]\boldsymbol{\mathcal{W}}+[\text{ad}_{\boldsymbol{\mathcal{V}}}]\boldsymbol{\mathcal{G}})\boldsymbol{\mathcal{L}}\boldsymbol{\mathcal{A}}\dot{\boldsymbol\theta} \\
\hfill{\boldsymbol{g}(\boldsymbol\theta)} &= \boldsymbol{\mathcal{A}}^T\boldsymbol{\mathcal{L}}^T\boldsymbol{\mathcal{G}}\boldsymbol{\mathcal{L}}\dot{\boldsymbol{\mathcal{V}}}_{\text{b}}\\
\hfill{\boldsymbol{J}^T(\boldsymbol\theta)} &= \boldsymbol{\mathcal{A}}^T\boldsymbol{\mathcal{L}}^T\\
\end{cases}
\end{equation}
%The meaning of the symbols will be explained in the appendix. 

For a flexible manipulator subject to tendon actuation the active joint torque is null, allowing to rewrite eq. \ref{dynamics} as:
%provided by the joint is constant to \(\boldsymbol{0}\), and it additionally receives the elastic torque and resistance torque from the flexible joint and the tension from the tendon. Therefore:
\begin{equation}
\begin{matrix}
\hspace{-20mm}\boldsymbol{0} =\boldsymbol{M}(\boldsymbol\theta)\ddot{\boldsymbol\theta}+\boldsymbol{c}(\boldsymbol\theta,\dot{\boldsymbol\theta})+\boldsymbol{g}(\boldsymbol\theta)\\
\hspace{10mm}+\boldsymbol{J}^T(\boldsymbol\theta)(\boldsymbol{\mathcal{F}}_{\text{e}}-\boldsymbol{\mathcal{F}}_{\text{t}}(\boldsymbol\theta,\boldsymbol{f}))+\boldsymbol{K}\boldsymbol\theta+\boldsymbol\mu\dot{\boldsymbol\theta}
\end{matrix}
\end{equation}
where \( \boldsymbol{K} \) and \( \boldsymbol\mu \) are the stacked vectors of the elastic and damping coefficients; The tendons forces stacked vector \(\boldsymbol{\mathcal{F}}_{\text{t}}(\boldsymbol\theta,\boldsymbol{f})\) is the stacked vectors of wrenches generated by the tension \( \boldsymbol{f} \) at each link provided by each tendon.

\subsection{Elastic-joint model}
For a general tendon-driven, flexible-joint manipulator with partially-constrained tendons, as defined in \cite{Rao2021}, each tendon passes through two consecutive cable-threads in each bead, as shown in the inset Fig. \ref{fig:side-by-side} and in Fig. \ref{fig:elastic}(a). This tendon arrangement, generates two forces applied to the bead on each side of the joint at the cable-thread locations. With reference to Fig. \ref{fig:elastic}, the model employed here assumes that:   %When considering two successive beads, which together constitute a joint,  we indicate the force applied on the preceding link the backward force \(\boldsymbol{f}_{\text{b}}\), and the one applied on the following link the forward force \(\boldsymbol{f}_{\text{f}}\). In the assumption that the cable-threads are located exactly halfway between two successive beads (or alternatively within the centre of rotation of the joint) then, because \(\boldsymbol{f}_{\text{b}}\) and \(\boldsymbol{f}_{\text{f}}\) are both applied on the straight line passing through the joint, we can further decompose their resultant force \(\boldsymbol{f}_{\text{r}}\) into two new forces at the position of each joint, along the direction of the link on both sides of the joint. 

% \begin{figure}[t!]
%     \centering
%     \includegraphics[width=\linewidth,trim= 10 10 10 10, clip]{figures/schematic_elastic_joint.png}
%     \caption{Schematic diagram of the Elastic joint model, depicting square beads connected by a flexible hinge.}
%     \label{fig:elastic}
% \end{figure}

\begin{enumerate}
    \item Fig. \ref{fig:elastic}(b): when a cable \(i\) is pulled, disregarding friction, the magnitude of the cable tension \(f_i\) is constant when passing through the beads and two actual forces \({\bm{f}_{\text{bt}}}_{ij}\) and \({\bm{f}_{\text{ft}}}_{ij}\) are generated at the two beads before and after each joint j. Here the subscripts of \({\bm{f}_{\text{bt}}}_{ij}\) indicate the tensile force generated by cable $i$ in the \textit{backward} link of joint $j$, or, in the case of \({\bm{f}_{\text{bt}}}_{ij}\) in the \textit{forward} link of joint $j$.
    
    \item Fig. \ref{fig:elastic}(c): if the spacing \(h\) between adjacent beads is small enough, \({\bm{f}_{\text{bt}}}_{ij}\) and \({\bm{f}_{\text{ft}}}_{ij}\) can be regarded as a force \({\bm{f}_{\text{t}}}_{ij}\) passing through the center of the elastic joint \(j\). The direction of cable tension is uniquely determined by the orientation of bead \(i\), so the direction vector \(\bm{n}_i\) of cable tension can be defined according to bead index \(i\). Hence, the relationship: \({\bm{f}_{\text{t}}}_{ij}=f_i(\bm{n}_{j-1}-\bm{n}_j)\) stands.
    
    \item Fig. \ref{fig:elastic}(d): the point of application of this force \({\bm{f}_{\text{t}}}_{ij}\) is at the intersection of the center lines of the two beads: when translated to the point of application, we indicate this force as \({\bm{f}_{\text{J}}}_{ij}\). Here, the force \({\bm{f}_{\text{J}}}_{ij}\) can be decomposed into two forces acting on each bead according to the above relationship, yielding: \({\bm{f}_{\text{bJ}}}_{ij}=f_i \bm{n}_{j-1}\) and \({\bm{f}_{\text{fJ}}}_{ij}=-f_i \bm{n}_j\).
    
    \item Fig. \ref{fig:elastic}(e): if the joint happens to be the last one in a segment, that is, the index of this joint is a multiple (using \(1\) as an example below) of the number of beads in each segment \(n_b\) , \({\bm{f}_{\text{J}}}_{ij}\) can still be decomposed, but the other end of bead \(n_b\) is subjected to the traction force \({\bm{f}_{\text{E}}}_{in_b}=f_i \bm{n}_{n_b-1}\) from the end of cable \(i\). At this time, \({\bm{f}_{\text{fJ}}}_{in_b}\) and \({\bm{f}_{\text{E}}}_{in_b}\) are equal in magnitude and opposite in direction, and the resultant torque becomes: 
    \begin{equation}\label{m_i}
        \bm{m}_i = \bm{r}_{i}\times \bm{n}_{n_b} f_i
        %\vspace{-5pt}
    \end{equation}
    where \(\bm{r}_i\) is the distance from the cable placement to the centroid line of the beads. 
    Since a cable has only one end, \(\bm{m}_i\) is unique for each index \(i\), consistent with the well known solution in \cite{Gravagne2003}. When bead \(n_b\) enters steady state, stiffness provides the torque to balance \(\bm{m}_i\), while the remaining force \({\bm{f}_{\text{bJ}}}_{in_b}\) in \({\bm{f}_{\text{J}}}_{in_b}\) can only act on the previous bead.
    \item Fig. \ref{fig:elastic}(f) When \({\bm{f}_{\text{J}}}_{ij}\) on each general joint is decomposed, it is found that for each beads j, \({\bm{f}_{\text{J}}}_{ij}\) will always cancel \({\bm{f}_{\text{J}}}_{i,j+1}\). This can be extended all the way to the first bead. Therefore, except for bead \(n_b\) receiving torque \(m_i\), the resultant force received by other beads from cables \(i\) is 0. 
\end{enumerate}
The above process allows calculation of the entire effect of tendon forces on the manipulator, but in order to adapt to screw-based formulation, a tendon force stacked matrix \(\boldsymbol{\mathcal{F}}_t\) consistent with the \(\boldsymbol{\mathcal{F}}_e\) (see Appendix) is introduced to facilitate the calculation. \(\boldsymbol{\mathcal{F}}_t\) is defined as the stack of the wrench \({\boldsymbol{\mathcal{F}}_t}_j\) exerted by the cables on each bead in its own frame. As discussed above, if the bead is not at the end of a segment, then \({\boldsymbol{\mathcal{F}}_t}_j\) is \(\bm{0}\); otherwise, \({\boldsymbol{\mathcal{F}}_t}_j\) is a pure force couple, which is the sum of all the force screws \(\boldsymbol{\mathcal{M}}_i\) of the cables connected to the bead, where the moment component of each \(\boldsymbol{\mathcal{M}}_i\) is \(\boldsymbol{m}_i\), and the force component is \(\bm{0}\). 
By introducing the stacked wrench of each segment \({\boldsymbol{\mathcal{F}}_{\text{tS}}}_i\), the above description is readily expressed as follows:
\vspace{0pt}
\begin{equation}
\setlength{\arraycolsep}{0pt}
\begin{matrix}
    \boldsymbol{\mathcal{F}}_{\text{t}} = \left[\begin{smallmatrix}
        {\boldsymbol{\mathcal{F}}_{\text{tS}}}_1\\
        {\boldsymbol{\mathcal{F}}_{\text{tS}}}_2\\
        \vdots\\
        {\boldsymbol{\mathcal{F}}_{\text{tS}}}_{n_s}\\
   \end{smallmatrix}\right]\in\mathbb{R}^{6n}\text{, }&
    {\boldsymbol{\mathcal{F}}_{\text{tS}}}_j = \left[\begin{smallmatrix}
    \hspace{-3mm}n_b-1\left\{\begin{smallmatrix}
        \mathbf{0}\\
        \vdots\\
        \mathbf{0}\\
        \end{smallmatrix}\right.\\
        \sum_{S_{j}}\boldsymbol{\mathcal{M}}_i
    \end{smallmatrix}\right]\in\mathbb{R}^{6n_b}\\
    \boldsymbol{\mathcal{M}}_i = \left[\begin{smallmatrix}
    \boldsymbol{m}_i\\
    0\\
    0\\
    0
    \end{smallmatrix}\right]\in\mathbb{R}^{6}\text{, }&
    \mathbf{0} = \left[\begin{matrix}
    0,0,0,0,0,0
    \end{matrix}\right]^T
\end{matrix}
%\vspace{-0.1cm}
\end{equation}
% \hl{(From this point on, this section is impossible to understand. We either remove it complitely or you try to explain it properly. In particular, eq. 12, 13 is not clear, is it really necessary?) Also eq. 15 appears out of nowhere and it is impossible to understand what it means.}

\noindent where \( \sum_{\small{S_j}}\boldsymbol{\mathcal{M}}_i \) is the sum of \( \boldsymbol{\mathcal{M}}_i \) on segment \(j\), \(\boldsymbol{m}_i\) here represents the component of \(\boldsymbol{m}_i\) in the bead's body frame and \(n_s\) is the number of segments. 
%Since each segment has the same structure, the format of the force screw on it is the same. Take a manipulator with four cables driving a segment as an example. According to the position of the cables relative to the centroid line, they can be numbered as 'x+', 'y+', 'x-', 'y-'. Among them, 'x+' is opposite to 'x-', and 'y+' is opposite to 'y-', so their screw can be expressed by a same vector with different signal.  Using eq.\ref{m_i}, the consistent “direction” part \(\boldsymbol{\mathcal{S}}_\mathcal{M}\)  can be separated from \( \sum_{\small{S_j}}\boldsymbol{\mathcal{M}}_i \). For the "magnitude" part, a coupling matrix \({\boldsymbol{\mathcal{P}}_\mathcal{M}}_j\) can select corresponding cable tension \(f_i\) from \(\bm{f}\). 
Upon rearrangement and inclusion of eq. \ref{m_i}, \( \sum_{\small{S_j}}\boldsymbol{\mathcal{M}}_i \) can be rearranged as follows for compactness:
% Having defined \( n_s \) as the number of segments and \( n_b \) as the number of beads in each segment, then the total number of joints is \( n = n_s \times n_b \). 
% Then
% \( \boldsymbol{T}_i = \boldsymbol{M}_i^{-1} \boldsymbol{M}_{J_{i+1}} \) represents the posture of link \( i \) in joint \( i+1 \)
% reference frame, where
% \( \boldsymbol{M}_{J_{i+1}} \) is the home configuration of joint \( i \) frame in the fixed reference frame. And \( \boldsymbol{M}_i \) is the home configuration of link  \( i \) frame in the fixed reference frame.
% %Because \( \boldsymbol{M}_i \) is a dual vector, which moves contrary to a vector when changing the axis, 
% \( {\text{Ad}}_{{\boldsymbol{T}_{i}}}^T \)  is the adjoint representation for \( \boldsymbol{M}_i \) to translate from link \( i \) axis to joint \( i+1 \)
% axis.
% Because \( \mathcal{M}_i \) can translate freely within the coordinate range of link. Therefore each non-zero term in  \( \boldsymbol{\mathcal{M}}_{\text{J}} \) can be written as:
%\setlength{\columnsep}{0.2pt}
%\setlength{\arraycolsep}{0.2pt}
\vspace{0pt}
\begin{equation}
\small{
\begin{array}{r@{\hskip 0pt}l}
\sum_{S_j}\boldsymbol{\mathcal{M}}_i  
&=\left[\begin{smallmatrix}\ \boldsymbol{r}_x\times\boldsymbol{n}_x&\boldsymbol{r}_y\times\boldsymbol{n}_y\\0&0\\0&0\\0&0\\\end{smallmatrix}\right]\left[\begin{smallmatrix}p_{xj1}&p_{xj2}&\cdots&p_{xjn}\\p_{yj1}&p_{yj2}&\cdots&p_{yjn}\\\end{smallmatrix}\right]\left[\begin{smallmatrix}f_1\\f_2\\\vdots\\f_n\\\end{smallmatrix}\right]\\
&=\boldsymbol{\mathcal{S}}_\mathcal{M}{{{\boldsymbol{\mathcal{P}}}_\mathcal{M}}_j}\boldsymbol{f}\\
\end{array}
}
\end{equation}
%\vspace{0pt}
\noindent with $\mathcal{S_{\mathcal{M}}}$ and $\mathcal{P}_{\mathcal{M}_j}$ respectively representing the first and second right hand side matrices in the above equation.

\section{Forward statics method}
%\setlength{\arraycolsep}{5pt}
%\subsubsection{tension input (FST)}
\subsection{Forward static with tendon force input}\label{FST}
Calculation of static configuration based on tendon force represents the conventional method in forward statics, extensively documented in the literature. Here, we define this as FST for short. At static equilibrium \( \dot{\boldsymbol\theta} \) and \( \ddot{\boldsymbol\theta} \) are null, simplifying the dynamics to
\begin{equation}
\boldsymbol{0} = \boldsymbol\tau(\boldsymbol\theta) = \boldsymbol{g}(\boldsymbol\theta)
+\boldsymbol{J}^T(\boldsymbol\theta)(\boldsymbol{\mathcal{F}}_{\text{e}}-\boldsymbol{\mathcal{F}}_{\text{t}}(\boldsymbol{f}))+\boldsymbol{K}\boldsymbol\theta
\end{equation}
which describes the joint torque $\bm{\tau}(\bm \theta)$ given a known force $\bm f$ at each tendon.
%This equation describes a function of \( \boldsymbol\tau(\boldsymbol\theta) \) concerning \( \boldsymbol\theta \) when the value of \( \boldsymbol{f} \) is given.
%When \( \boldsymbol\tau(\boldsymbol\theta) = \boldsymbol{0} \), the value of \( \boldsymbol\theta) \) is the true result corresponding to the certain \( f \). 
This can be conveniently solved iteratively by using the Newton-Raphson method by imposing a desired joint torque \(\boldsymbol\tau_D\) which for a static configuration must be \(\boldsymbol\tau_D=\boldsymbol{0}\): 
\begin{equation}\label{NR_FST}
\begin{cases}
\boldsymbol{\theta}_{\text{i}}&=\boldsymbol{\theta}_{\text{i}-1}+\alpha\left.(\frac{\boldsymbol\partial\boldsymbol\tau}{\boldsymbol\partial\boldsymbol\theta})^+\right|_{\text{i}-1}(\boldsymbol{\tau}_D\ -\ \boldsymbol{\tau}_{\text{i}-1})\\
\boldsymbol{\tau}_{\text{i}}&=\boldsymbol{\tau}(\boldsymbol{\theta}_{\text{i}})
\end{cases}
\end{equation}
where subscript $i$ indicates the  iteration  and \( (\cdot)^+\) is the pseudo-inverse operator. 

In eq. \ref{NR_FST}, the term $\partial \bm\tau/\partial \bm\theta$ requires clarification on its calculation process. By definition, rearranging eq. \ref{NR_FST} and differentiating w.r.t $\bm \theta$ yields, 
\begin{equation}
\small{
\frac{\boldsymbol\partial\boldsymbol\tau}{\boldsymbol\partial\boldsymbol\theta} = 
\frac{\boldsymbol\partial\boldsymbol{g}}{\boldsymbol\partial\boldsymbol\theta}+\frac{\boldsymbol\partial\boldsymbol{J}^T}{\boldsymbol\partial\boldsymbol\theta}(\boldsymbol{\mathcal{F}}_{\text{e}}-\boldsymbol{\mathcal{F}}_{\text{t}}(\boldsymbol{f}))
}
\end{equation}
where, using eq. \ref{Lterm} from appendix \ref{appendix_D} read as
\begin{equation}
\begin{cases}
\hfill\frac{\boldsymbol\partial\boldsymbol{g}}{\boldsymbol\partial\boldsymbol\theta} &= \footnotesize{\boldsymbol{\mathcal{A}}^T\boldsymbol\partial_{\boldsymbol\theta} \boldsymbol{\mathcal{L}}^T\boldsymbol{\mathcal{G}} \boldsymbol{\mathcal{L}}\dot{\boldsymbol{\mathcal{V}}}_{\text{b}}+}\\
 \rule{0.8em}{0pt} &  \rule{1em}{0pt}\footnotesize{\boldsymbol{\mathcal{A}}^T\boldsymbol{\mathcal{L}}^T\boldsymbol{\mathcal{G}} \boldsymbol\partial_{\boldsymbol\theta} \boldsymbol{\mathcal{L}}\dot{\boldsymbol{\mathcal{V}}}_{\text{b}}+
\boldsymbol{\mathcal{A}}^T\boldsymbol{\mathcal{L}}^T\boldsymbol{\mathcal{G}} \boldsymbol{\mathcal{L}}\boldsymbol\partial_{\boldsymbol\theta} \dot{\boldsymbol{\mathcal{V}}}_{\text{b}}
}\\
\hfill\frac{\boldsymbol\partial\boldsymbol{J}^T}{\boldsymbol\partial\boldsymbol\theta} &= \footnotesize{\boldsymbol{\mathcal{A}}^T\boldsymbol\partial_{\boldsymbol\theta} \boldsymbol{\mathcal{L}}^T
}\\
\end{cases}
\end{equation}
then, using the identity:
\begin{equation}
(\boldsymbol{I}-{\boldsymbol{\mathcal{W}}})\boldsymbol\partial_{\boldsymbol\theta}(\boldsymbol{I}-{\boldsymbol{\mathcal{W}}})^{-1} + \boldsymbol\partial_{\boldsymbol\theta}(\boldsymbol{I}-{\boldsymbol{\mathcal{W}}})(\boldsymbol{I}-{\boldsymbol{\mathcal{W}}})^{-1} = \boldsymbol{0}
\end{equation}
\noindent the term \( \partial_{\boldsymbol\theta} \boldsymbol{\mathcal{L}} \) can be calculated:
\begin{equation}
\partial_{\boldsymbol\theta} \boldsymbol{\mathcal{L}} = \partial_{\boldsymbol\theta}(\boldsymbol{I}-{\boldsymbol{\mathcal{W}}})^{-1}=
\boldsymbol{\mathcal{L}} \boldsymbol\partial_{\boldsymbol\theta} {\boldsymbol{\mathcal{W}}}\boldsymbol{\mathcal{L}}
\end{equation}
Calculation of the term $\partial_{\bm \theta}\bm {\mathcal{W}}$ requires that, inside the stacked matrix \(\boldsymbol{\mathcal{W}}\), we differentiate all the adjoint terms:
\vspace{-2pt}
\begin{equation}\begin{array}{rl}
\partial_{\boldsymbol\theta}\text{Ad}_{\boldsymbol{T}_{i,i-1}}& = 
\partial_{\theta_{i}}\text{Ad}_{(\boldsymbol{e^{-\boldsymbol{\mathcal{A}}_{i}\theta_i}M}_{i,i-1})}\otimes \boldsymbol{e}_{i}\\
&= 
\partial_{\theta_i}\text{Ad}_{\boldsymbol{e^{-\boldsymbol{\mathcal{A}}_{i}\theta_i}}}\text{Ad}_{\boldsymbol{M}_{i,i-1}}\otimes \boldsymbol{e}_{i}\\
&= 
\text{ad}_{\boldsymbol{(-\mathcal{A}}_{i})}\text{Ad}_{\boldsymbol{e^{-\boldsymbol{\mathcal{A}}_{i}\theta_i}}}\text{Ad}_{\boldsymbol{M}_{i,i-1}}\otimes \boldsymbol{e}_{i}\\
&=
- \text{ad}_{\boldsymbol{\mathcal{A}}_{i}}\text{Ad}_{\boldsymbol{T}_{i,i-1}}\otimes \boldsymbol{e}_{i}\\
\end{array}\end{equation}
We notice that \(\boldsymbol{\mathcal{W}}\) can be decomposed into an intermediate matrix \(\boldsymbol{\mathcal{W}}_i\) to better compute its partial derivative, i.e.:
\begin{equation}
\boldsymbol{\mathcal{W}}_i(\theta_i) = 
\left[\begin{smallmatrix}
\mathbf{0}&\cdots&\cdots&\cdots&\mathbf{0}&\mathbf{0}\\
\mathbf{0}&\cdots&\cdots&\cdots&\mathbf{0}&\mathbf{0}\\
\vdots&\ddots&\ddots&\ddots&\vdots&\vdots\\
\vdots&\ddots&\text{Ad}_{\boldsymbol{T}_{i,i-1}}(\theta_i)&\ddots&\vdots&\vdots\\
\vdots&\ddots&\ddots&\ddots&\vdots&\vdots\\
\mathbf{0}&\cdots&\cdots&\cdots&\mathbf{0}&\mathbf{0}\\\end{smallmatrix}\right]
\begin{array}{l}
\\
\\
\text{for } i \in [2,n]\\
\\
\text{  and }\boldsymbol{\mathcal{W}}_1 = \boldsymbol{0}
\end{array}
%\vspace{-5pt}
\end{equation}
and its partial derivative is:
\vspace{0pt}
\begin{equation}
\partial_{\theta_{i}}{\boldsymbol{\mathcal{W}}_{i}} = 
\left[\begin{smallmatrix}
\mathbf{0}&\cdots&\cdots&\cdots&\mathbf{0}&\mathbf{0}\\
\mathbf{0}&\cdots&\cdots&\cdots&\mathbf{0}&\mathbf{0}\\
\vdots&\ddots&\ddots&\ddots&\vdots&\vdots\\
\vdots&\ddots&- \text{ad}_{\boldsymbol{\mathcal{A}}_{i}}\text{Ad}_{\boldsymbol{T}_{i,i-1}}&\ddots&\vdots&\vdots\\
\vdots&\ddots&\ddots&\ddots&\vdots&\vdots\\
\mathbf{0}&\cdots&\cdots&\cdots&\mathbf{0}&\mathbf{0}\\\end{smallmatrix}\right]
\begin{array}{l}
\\
\\
\text{for } i \in [2,n]\\
\\
\text{and }\partial_{\theta_{1}}{\boldsymbol{\mathcal{W}}_{1}} = \boldsymbol{0}
\end{array}
%\vspace{-10pt}
\end{equation}
so that it can be re-written as:
\begin{equation}\label{partialtw}
\boldsymbol\partial_{\boldsymbol\theta} {\boldsymbol{\mathcal{W}}} = \boldsymbol\partial_{\boldsymbol\theta}\sum_i{\boldsymbol{\mathcal{W}}}_i = \sum_i\partial_{\theta_i}{\boldsymbol{\mathcal{W}}}_i \otimes \boldsymbol{e}_i
\vspace{-8pt}
\end{equation}
Now all the terms needed to compute eq.\ref{partialtw} in the definition of \(\frac{\boldsymbol\partial\boldsymbol{g}}{\boldsymbol\partial\boldsymbol\theta}\) and \(\frac{\boldsymbol\partial\boldsymbol{J}^T}{\boldsymbol\partial\boldsymbol\theta}\) have been made explicit.

\subsection{Forward static with tendon length input}\label{FSL}

The solution to the forward statics based on imposed tendon length starts from the definition of the tendon length model, $l_t$, following the classic approach of \cite{Murray1994}, as
\begin{equation}\label{tendon_length_model}
\boldsymbol{l}_{\text{t}}	= \boldsymbol{P}_{\theta}\boldsymbol\theta+\boldsymbol{l}_0-\boldsymbol\lambda_{\text{t}}\boldsymbol{f}
\end{equation}
\noindent where \( \boldsymbol{l}_0 \) the initial length of each tendon, and
\begin{equation}
\frac{\boldsymbol\partial\boldsymbol{l}_{\text{t}}}{\boldsymbol\partial\boldsymbol\theta}	= \boldsymbol{P}_{\theta}
\text{ and }
\frac{\boldsymbol\partial\boldsymbol{l}_{\text{t}}}{\boldsymbol\partial\boldsymbol{f}}	= -\boldsymbol\lambda_{\text{t}}
\end{equation}
\noindent here \( \boldsymbol{P}_{\theta} \) is a \textit{coupling matrix}
%this is composed of \( \pm{r} \) and \( 0 \) 
such that the term \( \boldsymbol{P}_{\theta}\boldsymbol\theta \) is the contribution of the joint changes to the cables lengths
%that \(r\) is the magnitude of \(\bm{r}_i\). 
while \( \boldsymbol\lambda_{\text{t}} \) is the stiffness of the tendon, and thus  \( \-\boldsymbol\lambda_{\text{t}}\boldsymbol{f} \) allows to account for tendon extensibility, taken here to be zero. 
% For the model verified later, the deformation of tendon is not considered, but \( \boldsymbol\lambda_{\text{t}}\) = 1e-16 is used to prevent singularities in the iteration.

Here, the solution of a forward static problem, which takes the tendon length as the input is demonstrated. This calculation approach is referred to as FSL and entails the calculation of a steady-state configuration of the manipulator under prescribed tendons' lengths \(\bm{l}_D\). This can be expressed by the simultaneous verification of the conditions \(\boldsymbol{\tau}(\bm \theta) = \boldsymbol{0}\) and \(\boldsymbol{l}_t = \boldsymbol{l}_D \), which can be solved by defining two augmented vectors:
\begin{equation}
\boldsymbol{X}_{\text{i}}=\left[\begin{matrix}\boldsymbol{\theta}_{\text{i}}\\\ \boldsymbol{f}_{\text{i}}\\\end{matrix}\right]
\text{ and}
\boldsymbol{Y}_{\text{i}}=\left[\begin{matrix}\boldsymbol{\tau}_{\text{i}}\\\ {\boldsymbol{l}_{t}}_{\text{i}}\\\end{matrix}\right]
\end{equation}
and employing the Newton-Raphson method, 
\begin{equation}\label{NR_length}
\begin{cases}
\hfill\boldsymbol{X}_{\text{i}} &= \boldsymbol{X}_{\text{i}-1}+\alpha\left.\frac{\boldsymbol\partial \boldsymbol{X}}{\boldsymbol\partial \boldsymbol{Y}}\right|_{\text{i}-1}(\boldsymbol{Y}_D\ -\ \boldsymbol{Y}_{\text{i}-1})\\
\hfill\boldsymbol{Y}_{\text{i}} &= \boldsymbol{Y}(\boldsymbol{X}_{\text{i}})
\end{cases}
\end{equation}
where:
\vspace{-10pt}
\renewcommand{\arraycolsep}{5pt}
\begin{equation}
\scalebox{1}{$\frac{\boldsymbol\partial \boldsymbol{X}}{\boldsymbol\partial \boldsymbol{Y}}$}=\scalebox{1}{${(\frac{\boldsymbol\partial \boldsymbol{Y}}{\boldsymbol\partial \boldsymbol{X}})}^+$}=
\left[\begin{matrix}\frac{\boldsymbol\partial\boldsymbol\tau}{\boldsymbol\partial\boldsymbol\theta}&\frac{\boldsymbol\partial\boldsymbol\tau}{\boldsymbol\partial \boldsymbol{f}}\\\frac{\boldsymbol\partial\boldsymbol{l}_{\text{t}}}{\boldsymbol\partial\boldsymbol\theta}&\frac{\boldsymbol\partial\boldsymbol{l}_{\text{t}}}{\boldsymbol\partial \boldsymbol{f}}\\\end{matrix}\right]^+
\end{equation}
In \(\frac{\boldsymbol\partial\boldsymbol\tau}{\boldsymbol\partial \boldsymbol{f}}\), the calculation of \(\frac{\boldsymbol\partial\boldsymbol{\mathcal{F}}_{\text{t}}}{\boldsymbol\partial\boldsymbol{f}}\) is also required: 
% The variables in eq. \ref{FeqM} upon differentiation by the cable tension can be calculated as follows. 
\begin{equation}
\begin{matrix}
\frac{\boldsymbol\partial\boldsymbol{\mathcal{F}}_{\text{t}}}{\boldsymbol\partial\boldsymbol{f}}=\left[\begin{smallmatrix}
        \boldsymbol\partial_{\boldsymbol{f}}{\boldsymbol{\mathcal{F}}_{\text{tS}}}_1\\
        \boldsymbol\partial_{\boldsymbol{f}}{\boldsymbol{\mathcal{F}}_{\text{tS}}}_2\\
        \vdots\\
        \boldsymbol\partial_{\boldsymbol{f}}{\boldsymbol{\mathcal{F}}_{\text{tS}}}_{n_s}\\
   \end{smallmatrix}\right],
   \boldsymbol\partial_{\boldsymbol{f}}{\boldsymbol{\mathcal{F}}_{\text{tS}}}_j = \left[\begin{smallmatrix}
    \hspace{-1mm}n_b-1\left\{\begin{smallmatrix}
        \mathbf{0}\\
        \vdots\\
        \mathbf{0}\\
        \end{smallmatrix}\right.\\
        \boldsymbol{\mathcal{S}}_\mathcal{M}{{{\boldsymbol{\mathcal{P}}}_\mathcal{M}}_j}
    \end{smallmatrix}\right]
   \end{matrix}
   \end{equation}
%Since \(\boldsymbol{f}\) changes along the gradient direction of \(\boldsymbol{Y}(\boldsymbol{\theta},\boldsymbol{f})\) during iteration, 
This approach suffers from the fact that negative values of $\bm f$ can occur, which would imply that tendon forces act by pushing instead of pulling. To prevent this unrealistic behavior, we introduce an independent variable \(\bm{u}\) and a kernel function \(f_j(u_j)\) that is monotonic and smooth and acts by constraining the value of \(f_j\) such that \(f_j =0\) when \(u_j<0\), and \(f_j = u_j\) when \(u_j>0\). A candidate function for this is:
\begin{equation*}
f_j(u_j)=\frac{1}{2}(\sqrt{4c+{u_j}^2}+u_j)
\end{equation*}
This correction can be implemented in the solution process of eq. \ref{NR_length} by modifying the vector $\bm{X}_i$ such that:
\noindent\begin{equation}
{\boldsymbol{X}}_{\text{i}} \rightarrow\tilde{\boldsymbol{X}}_{\text{i}}=\left[\begin{matrix}\boldsymbol{\theta}_{\text{i}}\\\ \boldsymbol{u}_{\text{i}}\\\end{matrix}\right],
% \end{equation}
% \begin{equation}
\scalebox{1}{$\frac{\boldsymbol\partial \boldsymbol{X}}{\boldsymbol\partial \boldsymbol{Y}}$}\rightarrow\scalebox{1}{$\frac{\boldsymbol\partial \tilde{\boldsymbol{X}}}{\boldsymbol\partial \boldsymbol{Y}}$}=
\left[\begin{matrix}\frac{\boldsymbol\partial\boldsymbol\tau}{\boldsymbol\partial\boldsymbol\theta}&\frac{\boldsymbol\partial\boldsymbol\tau}{\boldsymbol\partial \boldsymbol{f}}\frac{\boldsymbol\partial\boldsymbol{f}}{\boldsymbol\partial\boldsymbol{u}}\\
\frac{\boldsymbol\partial\boldsymbol{l}_{\text{t}}}{\boldsymbol\partial\boldsymbol\theta}&\frac{\boldsymbol\partial\boldsymbol{l}_{\text{t}}}{\boldsymbol\partial \boldsymbol{f}}\frac{\boldsymbol\partial\boldsymbol{f}}{\boldsymbol\partial\boldsymbol{u}}\\\end{matrix}\right]^+
\end{equation}
with:
\begin{equation}
\scalebox{1.2}{$\frac{\boldsymbol\partial \boldsymbol{f}}{\boldsymbol\partial \boldsymbol{u}}$}=\text{diag}(\frac{{u_1}^2}{{u_1}^2+c},\cdots,\frac{{u_{n_l}}^2}{{u_{n_l}}^2+c})
\end{equation}
where \({n_l}\) is the total number of tendons.
\section{Experimental Validation}
\begin{figure}[t]
    \centering
   \includegraphics[width=0.95\columnwidth]{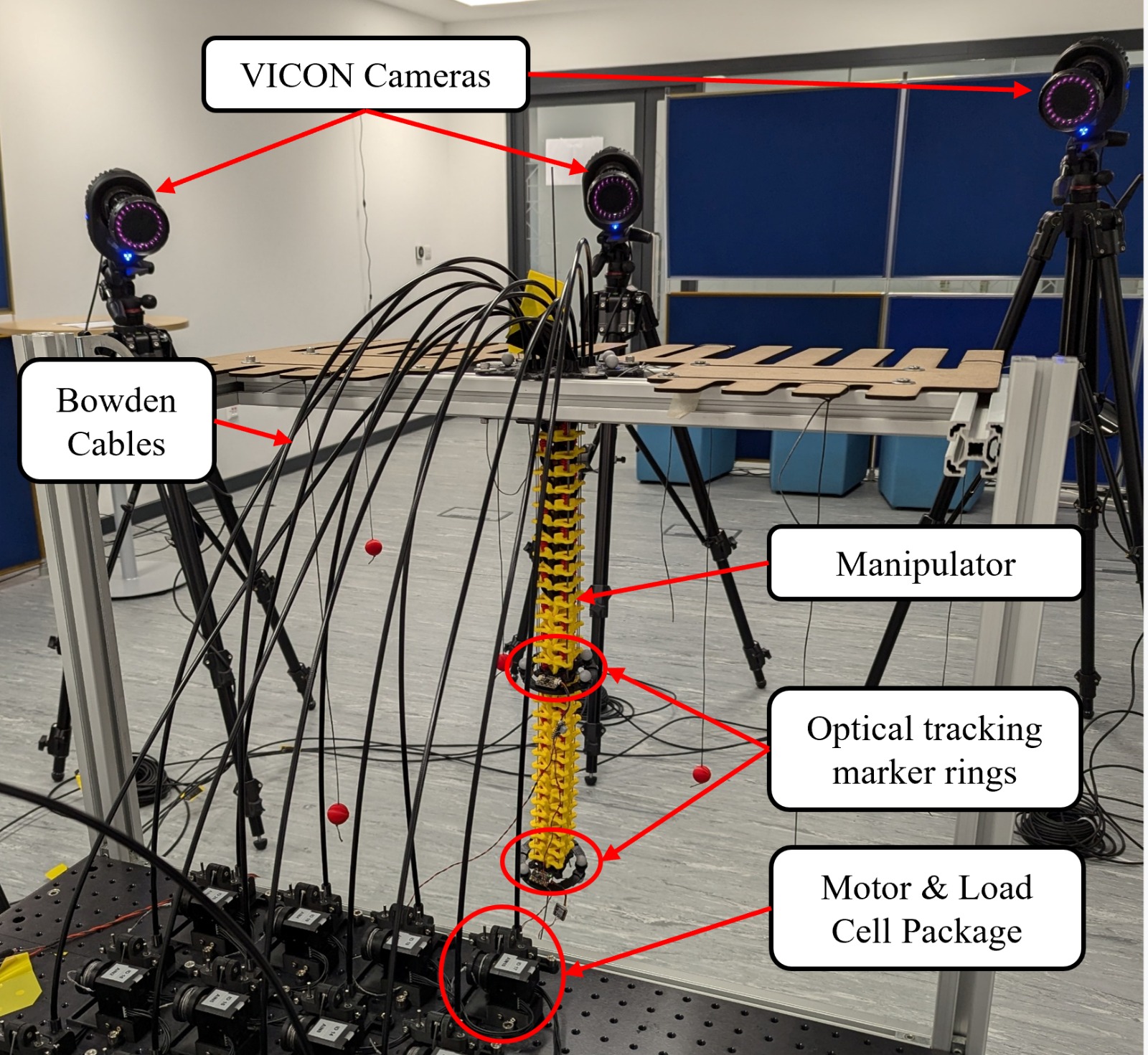}
    \caption{The experimental setup used to extract the posture data and cable tension data for evaluating the model.}\vspace{-0.5cm}
    \label{experimental_setup}
\end{figure}
To validate the effectiveness of the model in representing the behaviour of a real system, an experimental study was undertaken using a similar platform to that described in \cite{Walker2024, Walker2025}. The manipulator has a modular structure and is constructed of identical overlapping repeating pieces, with successive beads rotating around respectively orthogonal axes; see Fig. \ref{fig:side-by-side}. Each bead is 29.5mm in height, 62mm wide and weighs 10g, with 1 segment of the manipulator comprising 16 beads (15 regular beads and 1 with the addition of a holster for an IMU). Concurrent beads are held together with compressible, elastic TPU hinges and a 1.5mm diameter NiTi rod runs down the centre of the manipulator cross-section, equal in length to the desired length of the manipulator ($\approx$ 0.7m in this work). Actuation is performed using a tendon-driven approach, with 4 tendons used for each of the two segments. These are routed to a set of motors via bowden cables; see Fig. \ref{experimental_setup}.
\subsection{Measurement}
For these experiments, the manipulator has two independently actuated segments and is operated in air, using a VICON motion capture system to track the position in Cartesian space and orientation in quaternions of each segment tip, also shown in Fig. \ref{experimental_setup}. Positional measurements can be used directly for validation with respect to the manipulator base, whereas the quaternion measurements %$[x_q, y_q, z_q, w_q]^{T}$ 
are transformed into the configuration variables required for the PCC model forward kinematics. 
% according to:
% \begin{equation}
%     \phi_{Ci} = \arctan\left( \frac{x_q z_q-w_q y_q}{y_q z_q+w_q x_q}\right)
% \end{equation}
% \begin{equation}
%     \theta_{Ci} = \arccos(2w_q^2-1+2z_q^2)
% \end{equation}

Tensions measurements are obtained by integrating a Dynamixel servo motor (XM430-W350), a load cell (TE Connectivity, FC2231), and three pulleys for each cable. %The tension of the cable will be used in the calculation of FST. 
The pulleys ensure that the forces acting along the cable remain co-planar, so that as the cable experiences tension, it produces downward force and compresses the load  \cite{Walker2025}.

Static configurations suitable for validation of the proposed solution methodologies are selected based on the end-effector velocity and acceleration not exceeding 0.5 mm/s and 25 mm/\(\text{s}^\text{2}\) respectively and the cable tensions change rate being less than 0.1 N/s, representative of static configurations. This yields five suitable configurations, three of which are used for validation of the FST and the remaining two for the FLS. Ten further instances are used for evaluation of the model performance as the configurations deviates from a constant curvature regime. 
%After collecting the data, due to the limitations of statics and in order to facilitate the comparison between FST and PCC, data screening is also required. The selected conditions are that the end-effector velocity and acceleration do not exceed 0.5 mm/s and 25 mm/\(\text{s}^\text{2}\) respectively, the cable tensions change rate is less than 0.1 N/s, and the duration is at least 0.1 s.
% \begin{figure}
%     \centering
%     \includegraphics[width=\linewidth]{figures/grouped.png}
%     \caption{The manipulator used within the experiments to validate the model, showing (a) the discretised motions between successive beads and (b) the physical manipulator in a curved and straight posture.}
%     \label{manipulator}
% \end{figure}
\subsection{Validation of forward statics with tendons force input}
\begin{figure}[b!]
    \centering
    \includegraphics[width=0.48\textwidth,trim= 40 20 20 60, clip]{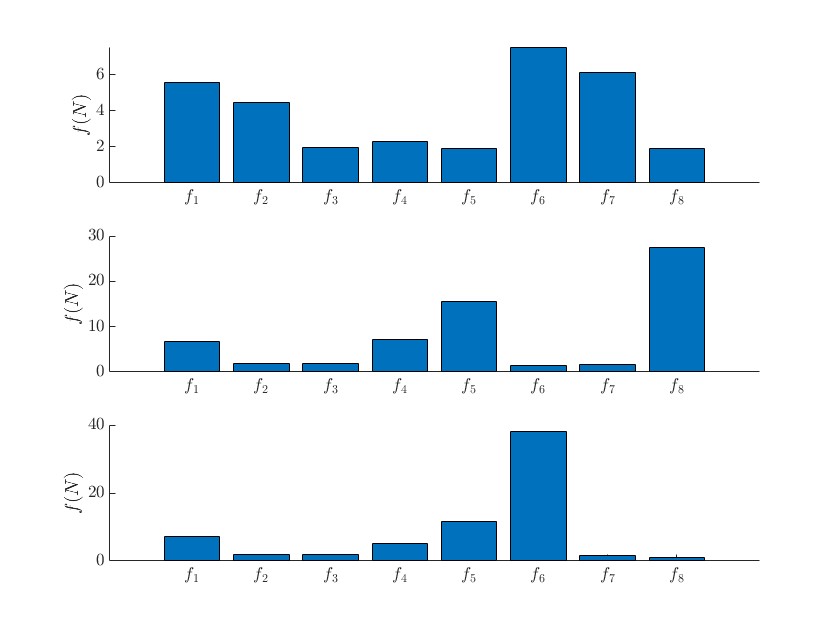}
    \caption{Experimentally measured tension on each of the 8 tendons for the three configurations tested. These are the input to the FST. }
    \label{force_exp}
\end{figure}
\begin{figure}[t!]
    \centering
    \includegraphics[width=0.48\textwidth,trim= 20 20 30 0, clip]{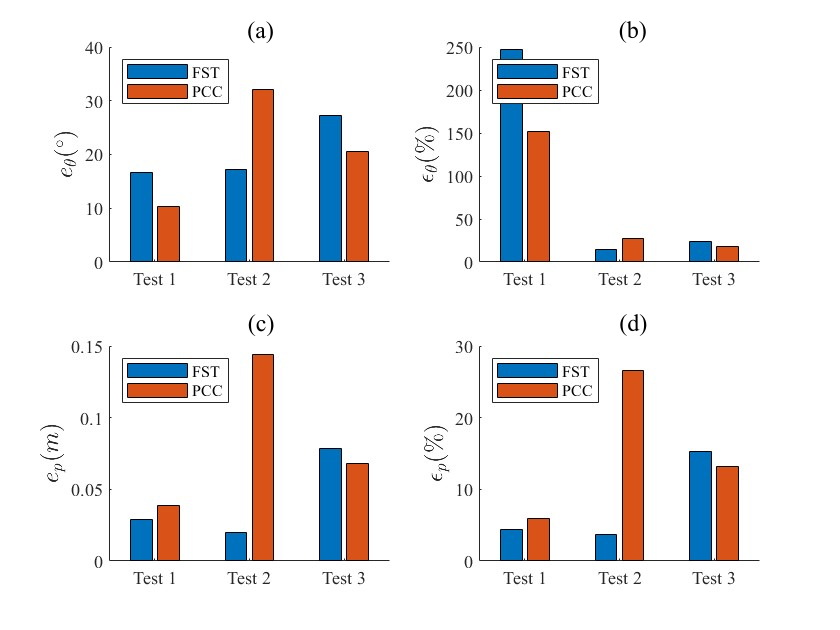}
    \caption{Error in the estimation of the end-effector orientation (a), (b) and position (c), (d) when estimated through the FST model and the PCC kinematic model.}
    \label{errors}
    \vspace{-0.3cm}
\end{figure}

\begin{figure}[t!]
    \centering
  \includegraphics[width=0.48\textwidth,trim= 20 20 20 0, clip]{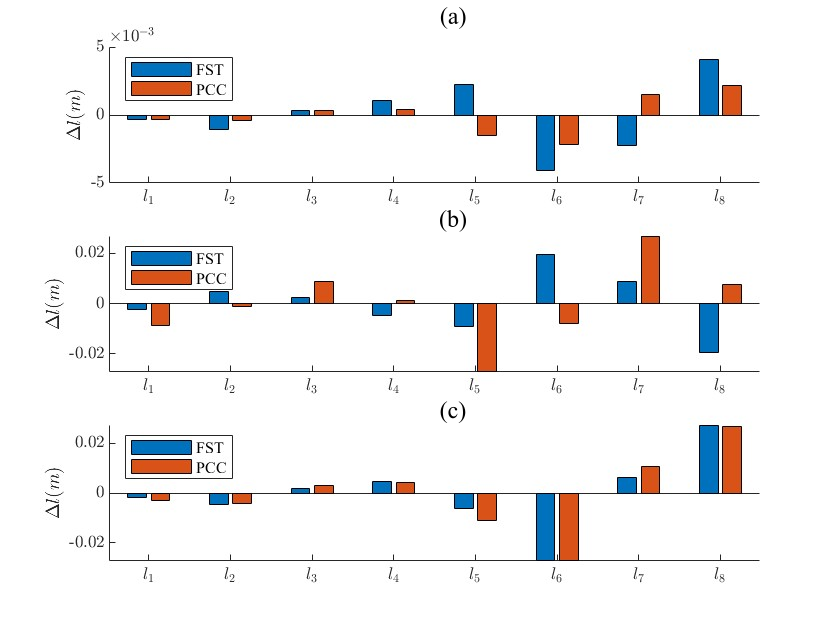}
    \caption{Results of the estimation of tendon length change in the three configurations tested (a), (b) and (c). Comparison between the present model and purely kinematic calculation from a PCC model.}\vspace{-0.5cm}
    \label{model_vs_pcc_cable_length}
\end{figure}
% \begin{figure}[t!]
%     \centering
%     \includegraphics[width=0.48\textwidth]{figures/c1-3r.jpg}
%     \caption{Three configurations estimated based on the cable tension input from experiments.}
%     \label{art_schematic}
% \end{figure}

% \begin{figure}[t!]
%     \centering
%     \includegraphics[width=0.48\textwidth]{figures/postures_test.png}
%     \caption{Various configurations estimated by the model based on a set of cable tension inputs extracted during the experiments, showing (a) **insert parameters**, (a) **insert parameters**, (a) **insert parameters**, and (d) a top down view of **insert parameters**.}
%     \label{postures}
%\end{figure}
To verify the FST and the associated solution method, the measured cable tensions are directly used as the input for the solution of static manipulator postures. Fig. \ref{force_exp} shows the cable tensions in three steady-state intervals. %uses \(\phi_{Ci}\) and \(\theta_{Ci}\) as input to obtain PCC's cables' lengths, posture of end-effector and configuration.
  %Fig. \ref{errors} shows the steady-state configurations obtained by FST in these three intervals. 
Using these tensions as the input $\bm f$ to the FST as per section \ref{FST}, yields three static postures: the predicted configurations are compared against experiments in Fig. \ref{errors}, which shows the absolute orientation error \(e_\theta\), relative orientation error \(\epsilon_\theta\), absolute position error \(e_p\) and relative position error \(\epsilon_p\) of the end effector in these three intervals (see definition in appendix \ref{appendix_E} ). For comparison, the PCC kinematic solution is also used as a reference to the accuracy of the FST in Fig. \ref{errors}. 
%Fig. \ref{model_vs_pcc_cable_length} shows the cable lengths calculated by PCC and FST for each interval respectively.
% \begin{figure*}[h!]
%     \centering
%     \begin{subfigure}{0.45\textwidth}
%         \centering
%         \includegraphics[width=\linewidth,trim= 600 120 60 120, clip]{figures/ISL2.png}
%     \end{subfigure}
%     \vrule width 0.2pt
%     \begin{subfigure}{0.45\textwidth}
%         \centering
%         \includegraphics[width=\linewidth,trim= 600 120 60 120, clip]{figures/ISL3.png}
%     \end{subfigure}

%     % \vspace{10pt}
%     \hrule height 0.2pt
%     % \vspace{10pt}

%     \begin{subfigure}{0.45\textwidth}
%         \centering
%         \includegraphics[width=\linewidth,trim= 600 117 60 120, clip]{figures/r-FSL2.png}
%     \end{subfigure}
%     \vrule width 0.2pt
%     \begin{subfigure}{0.45\textwidth}
%         \centering
%         \includegraphics[width=\linewidth,trim= 600 117 60 120, clip]{figures/r-FSL3.png}
%     \end{subfigure}

%     % \vspace{10pt}
%     \hrule height 0.2pt
%     % \vspace{10pt}
    
%     \begin{subfigure}{0.45\textwidth}
%         \centering
%         \includegraphics[width=\linewidth,trim= 600 125 60 180, clip]{figures/r-PCC2.png}
%     \end{subfigure}
%     \vrule width 0.2pt
%     \begin{subfigure}{0.45\textwidth}
%         \centering
%         \includegraphics[width=\linewidth,trim= 600 125 60 180, clip]{figures/r-PCC3.png}
%     \end{subfigure}
\begin{figure*}
    \centering
    \includegraphics[width=0.98\textwidth,trim= 0 0 0 0, clip]{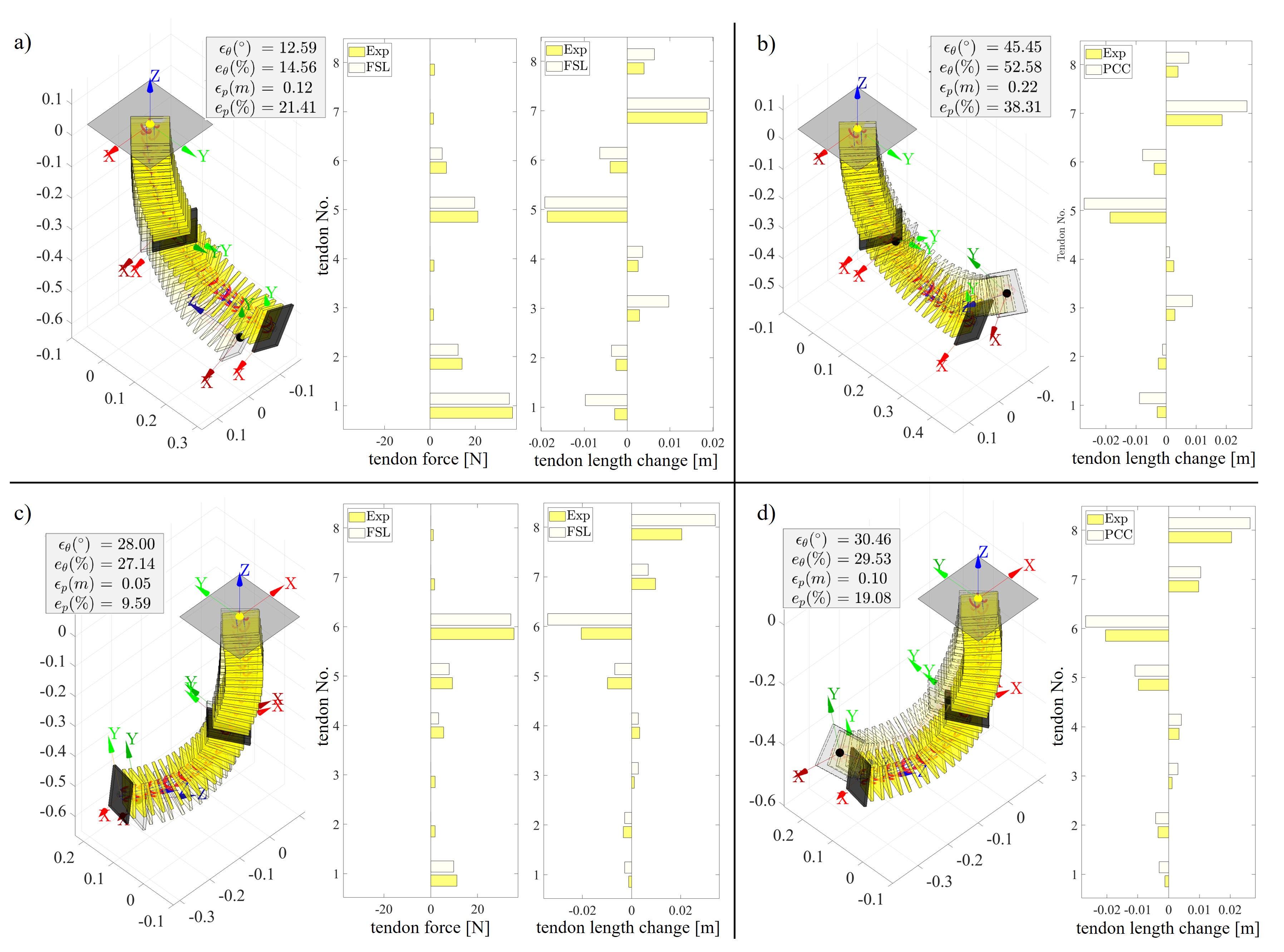}
    \caption{Results from forward static with tendon-length input in two configurations (a, b) and (c, d): in each figure the experimental data is colored in yellow and the simulated ones in white; the insets in each figure report absolute and relative error of the end-effector orientation and position with respect to experiments. In (a) and (c) the results from the proposed FSL solution method of section \ref{FSL} are shown, demonstrating both tendon length and tendon force estimation. In (b) and (d) the estimates from PCC model. }\vspace{-0.5cm}
    \label{fig.FSL}
\end{figure*}
%\(\bm{R}_e\), \(\bm{R}_t\), and \(\bm{R}_0\) are the  posture of the end effector from experiment, the posture of the end effector calculated by the model, and the posture of the fixed frame, respectively. \(\bm{p}_e\), \(\bm{p}_t\), and \(\bm{p}_0\) are the positions of these frame.
The results show that, like PCC, FST has large errors in the end-effector orientation, especially for small absolute angles. However, FST has a higher positional accuracy than PCC, especially over larger displacements as supported by the data in Fig. \ref{errors}(c)-(d). This shows the potential of the model to work well in the task-space. In addition, solving for eq. \ref{tendon_length_model} in the three configurations tested, provides a way to compare tendon-length estimates from the current model and the traditional kinematic PCC approach of \cite{ Rucker2011, Rao2021}, as shown in Fig. \ref{model_vs_pcc_cable_length}.  

\subsection{Validation of forward statics with tendons length input}

Finally we use the two remaining experimental tests to assess the effectiveness of the method presented in section \ref{FSL}. These two configurations are chosen for their pronounced deviation from a PCC approximation, so to evaluate the robustness of the proposed method in such conditions. 

\begin{figure}
\centering
\includegraphics[width=1.\columnwidth,trim= 22 15 37 20, clip]{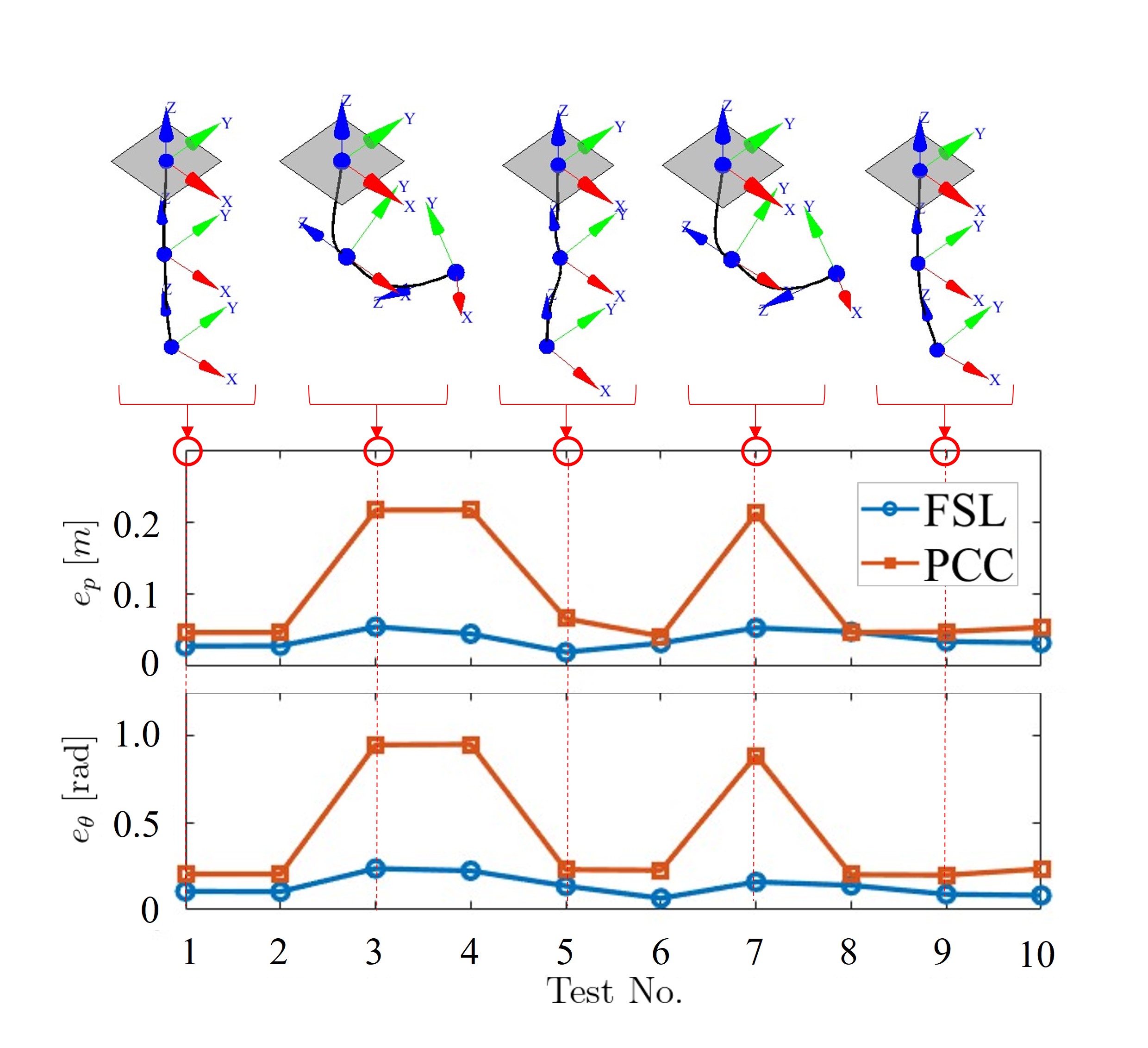}
\caption{Absolute positional and orientation error estimate from FSL and PCC model across ten configurations. In the top side of the figure, configurations 1, 3, 5, 7 and 9 are shown for readers' convenience.}
\label{extratests}
\end{figure}

In Fig. \ref{fig.FSL}, two configurations are shown: Fig. \ref{fig.FSL}(a, b) and Fig. \ref{fig.FSL}(c, d) with the experimental case coloured yellow and the simulated ones, either from FSL or PCC, coloured in white. In \ref{fig.FSL}(a) and (c), the configurations obtained with the FSL method are shown, where the tendon lengths are taken as the input to the forward statics problem to produce estimates of both configurations and tendon forces. Fig. \ref{fig.FSL}(b) and (d) depicts the configurations obtained from the PCC model and the associated tendon length estimate.
Orientation and positional errors of the end-effector w.r.t the experimental dataset are also reported in each figures' inset. These results confirm the effectiveness of the FSL method to produce an accurate solution to the forward statics when starting from a given tendon-length; they also show improved capability to predict the end-effector position outside of the PCC approximations margins. Notably, the proposed model also produces very accurate estimations of the tendon forces as the by-product of the FSL solution.  
%\textcolor{red}{With regards to computation, for the manipulator used in our experiments (excluding unrelated code such as animation generation) the FSL method typically achieves convergence with a residual below 1e-8 within 3 seconds. While this speed may not be considered fast, it is acceptable for open-loop control. Moreover, we have not yet optimized the code for execution speed, for example rewriting the Newton-Raphson loop in C++. In this regard, there is still considerable room for improvement in our method.}
Finally, the tendon length information, alone, is further used to estimate end-effector orientation and position across ten additional configurations, Fig. \ref{extratests}, demonstrating overall FSL accuracy outside the PCC regime.
%In order to test the correctness of the FSL, it is necessary to measure the changes of the cables lengths. However, current equipment does not support direct measurement of this, so a tortuous method is proposed. In the first step, using measured tendon forces as input to calculate the cables’ length by FST. As mentioned above, the results of FST have higher accuracy when the displacement is large, so it can be the cables' lengths for FSL validation, instead of the actual measurement value.
%In the second step, perform FSL, taking the cables’ lengths obtained in the first step as input to calculate a  group of tendon forces, the posture of end-effector and the configuration. 
%The third step is to use \(\phi_{Ci}\) and \(\theta_{Ci}\) to do PCC and obtain another group of the posture of end-effector and the configuration. Now we can compare these two groups of data with the experiment result.
%In addition, in order to facilitate comparison with the actual posture of end-effector from experiment, this paper uses the inverse statics method to generate the posture corresponding to the posture and obtains another set of cables’ length. Since this article does not involve the explanation of inverse statics, only its configuration is used as the real manipulator's for comparison.
%The comparison is shown in the Fig. \ref{fig.FSL} . It can be observed that the tendon forces and tendon lengths of FSL are very close to the experimental results and PCC, and compared with PCC, the configuration of FSL is closer to the experimental results.

\section{Conclusions}

% \begin{table*}[t]
%     \centering
%     \renewcommand{\arraystretch}{1.5} % 增加表格行距
%     \setlength{\tabcolsep}{5pt} % 调整列间距
%     \begin{tabular}{|c|c|c|}
%         \hline
%         % \includegraphics[width=0.33\textwidth,trim= 650 125 420 130, clip]{figures/ISL1.png} & 
%         \includegraphics[width=0.48\textwidth,trim= 600 125 360 120, clip]{figures/ISL2.png} & 
%         \includegraphics[width=0.48\textwidth,trim= 600 125 360 120, clip]{figures/ISL3.png} \\ 
%         \hline
%         % \includegraphics[width=0.33\textwidth,trim= 520 60 330 100, clip]{figures/FSL1.png} & 
%         \includegraphics[width=0.48\textwidth,trim= 650 125 420 130, clip]{figures/FSL2.png} & 
%         \includegraphics[width=0.48\textwidth,trim= 650 125 420 130, clip]{figures/FSL3.png} \\ 
%         \hline
%         % \includegraphics[width=0.33\textwidth,trim= 520 60 330 100, clip]{figures/PCC1d.png} & 
%         \includegraphics[width=0.48\textwidth,trim= 600 125 360 180, clip]{figures/PCC2d.png} & 
%         \includegraphics[width=0.48\textwidth,trim= 600 125 360 180, clip]{figures/PCC3d.png} \\ 
%         \hline
%     \end{tabular}
%     \caption{Comparison of FSL and PCC results}
%     \label{tab:images}
% \end{table*}
This paper presents a screw-based modeling methodology aimed at addressing the open-loop forward statics challenge associated with tendon-driven hyper-redundant flexible-joint manipulators. The introduced technique permits both tendon length and tendon force as input variables for the solution of forward statics, thereby facilitating forecasting of the manipulator’s static configuration while accounting for gravitational influence, distributed stiffness, and non-uniform tendon routing. The results show improved accuracy over conventional piecewise constant curvature models. More importantly, by obviating the need for tendon force as the input to the statics solution, the proposed approach is promising in offering a direct way to perform static control by using only tendon-length measurements, potentially bypassing the need for more advanced sensors or external visual feedback.

While the model effectively handles segment-wise stiffness variations, its applicability to highly nonlinear elastic behaviors requires further investigation. More importantly, this study concerns itself with forward statics, leaving room for future work towards the solution of inverse statics with tendon-length input and extending the approach to forward and inverse dynamics.

%Future research could explore several directions to further enhance and expand this work. First, improving tendon force modeling to account for static friction and hysteresis could enhance the model’s robustness. Second, eliminating the statics assumption and proposing tendon length driven dynamics. Finally, investigating its application in motion control and path planning could contribute to the advancement of hyper-redundant tendon-driven manipulators in practical robotic applications
%%%%%%%%%%%%%%%%%%%%%%%%%%%%%%%%%%%%%%%%%%%%%%%%%%%%%%%%%%%%%%%%%%%%%%%%%%%%%%%%
\section*{APPENDIX}
\subsection{Anti-dual Form in \(\mathbb{R}^3\)}\label{appendix_A}
For a vector, which can be seen as a 1-form written as:
\begin{equation}
\boldsymbol{v} = [v_1,v_2,v_1]^T \in \Omega^1(\mathbb{R}^3)
% \begin{bmatrix}
% v_1\\
% v_2\\
% v_3
% \end{bmatrix} \in \Omega^1(\mathbb{R}^3)
\end{equation}
The anti-dual of a 1-form can be written as:
\renewcommand{\arraycolsep}{5pt}
\begin{equation}
{}^{\ast}\boldsymbol{v} = -{}^{\star}\boldsymbol{v} = 
\begin{bmatrix}
\rule{0.8em}{0pt}0 & -v_3 & \rule{0.8em}{0pt}v_2\\
\rule{0.8em}{0pt}v_3 & \rule{0.8em}{0pt}0 & -v_1\\
-v_2 & \rule{0.8em}{0pt}v_1 & \rule{0.8em}{0pt}0
\end{bmatrix} \in \Omega^2(\mathbb{R}^3)
\end{equation}
where \( \star \) is the Hodge star operator.
\subsection{Adjoint Representation of \(\mathfrak{se}(3)\) and \(SE(3)\)}\label{appendix_B}
For a twist $\bm{\mathcal{V}}$ 
%\begin{equation}
% \boldsymbol{\mathcal{V}} = 
% \begin{bmatrix}
% \boldsymbol{\omega}\\
% \boldsymbol{v} 
% \end{bmatrix}
% \end{equation}
the operator $\text{ad}_{\mathcal{\bm V}}$, commonly referred to as the Lie bracket, is defined as 
\begin{equation}
\text{ad}_{\boldsymbol{\mathcal{V}}} =
\begin{bmatrix}
{}^{\ast}\boldsymbol{\omega} & {}^{\ast}\boldsymbol{v} \\
\boldsymbol{0} & {}^{\ast}\boldsymbol{\omega}
\end{bmatrix}.
\end{equation}
and for a generic homogenous tranformatin matrix $\bm T$, the adjoint operator $\text{Ad}_{\bm T}$ is
% \begin{equation}
% \boldsymbol{T} =
% \begin{bmatrix}
% \boldsymbol{R} & \boldsymbol{p} \\
% \boldsymbol{0} & 1
% \end{bmatrix}
% \end{equation}
\begin{equation}
\text{Ad}_{\boldsymbol{T}} =
\begin{bmatrix}
\boldsymbol{R} & {}^{\ast}\boldsymbol{p}\boldsymbol{R} \\
\boldsymbol{0} & \boldsymbol{R}
\end{bmatrix}.
\end{equation}
\renewcommand{\arraycolsep}{0pt}
% \subsection{Moment of Inertia and Coordinate Transformation of Spatial Inertia}\label{appendix_C}
% The moment of inertia of a rigid body \(a\) can be written as:
% \begin{equation}
% \boldsymbol{\mathcal{I}} = 
% \begin{bmatrix}
% \mathcal{I}_{11} & \mathcal{I}_{12} & \mathcal{I}_{13}\\
% \mathcal{I}_{12} & \mathcal{I}_{22} & \mathcal{I}_{23}\\
% \mathcal{I}_{13} & \mathcal{I}_{32} & \mathcal{I}_{33}
% \end{bmatrix}
% \end{equation}
% where:
% \begin{equation}
% \begin{cases}
% \mathcal{I}_{11} = \int_{a}\left(y^2+z^2\right)\rho\left(x,y,z\right)dV\\
% \mathcal{I}_{22} = \int_{a}\left(z^2+x^2\right)\rho\left(x,y,z\right)dV\\
% \mathcal{I}_{33} = \int_{a}\left(x^2+y^2\right)\rho\left(x,y,z\right)dV\\
% \mathcal{I}_{12} = \mathcal{I}_{21} = -\int_{a} x y\rho\left(x,y,z\right)dV\\
% \mathcal{I}_{23} = \mathcal{I}_{32} = -\int_{a} y z\rho\left(x,y,z\right)dV\\
% \mathcal{I}_{31} = \mathcal{I}_{13} = -\int_{a} z x\rho\left(x,y,z\right)dV\\
% \end{cases}
% \end{equation}
% The symbol \( \int_{a}(\cdot)dV \) represents the integration of all volumes of rigid body \(a\). The coordinate system used for integration is consistent with the frame, where the inertia is.
% A spatial inertia expressed in frame \(\{a\}\) can be transformed to other frame \(\{b\}\) as follows:
% \begin{equation}
% \boldsymbol{\mathcal{G}}_b = 
% \text{Ad}_{\boldsymbol{T_{ab}}}^T\boldsymbol{\mathcal{G}}_a\text{Ad}_{\boldsymbol{T_{ab}}}
% \end{equation}

\subsection{Explanation of Symbols in Robotics Dynamics}\label{appendix_D}
%Create a reference frame \(\{x^j_L\}_i\) on each link \(i\) in turn, 
Given an arbitrary reference frame attached to a link $i$, identified by $T_i$, then the twist of such link can be expressed as \( {\boldsymbol{\mathcal{V}}}_i\) and its wrench \( {\boldsymbol{\mathcal{F}}}_i\), then we define a vector of twists and wrench of all link as follows:
\begin{equation}
{\boldsymbol{\mathcal{V}}}=\left[\begin{matrix}{\boldsymbol{\mathcal{V}}}_1\\{\boldsymbol{\mathcal{V}}}_2\\\vdots\\{\boldsymbol{\mathcal{V}}}_n\\\end{matrix}\right]\in\mathbb{R}^{6n}, {\boldsymbol{\mathcal{F}}}=\left[\begin{matrix}{\boldsymbol{\mathcal{F}}}_1\\{\boldsymbol{\mathcal{F}}}_2\\\vdots\\{\boldsymbol{\mathcal{F}}}_n\\\end{matrix}\right]\in\mathbb{R}^{6n}
\end{equation}
%The speed and acceleration of the rack have not yet been defined. And the force rotation of the end effector. 
% Given the twist and acceleration of the base frame $T_0$ as \({\boldsymbol{\mathcal{V}}}_0,{\dot{\boldsymbol{\mathcal{V}}}}_0\) and the wrench at the end-effector as   
% \(\boldsymbol{\mathcal{F}}_{n+1}\) then then base twist and acceleration expressed w.r.t link 1 can be achieved as
% %Use adjoint representation to transform them to the closest frame in \(\{x^j_L\}_i\) and stack them up:
% \begin{equation}
% {\boldsymbol{\mathcal{V}}}_{b}=\left[\begin{matrix}{\text{Ad}}_{{\boldsymbol{T}}_{10}}{\boldsymbol{\mathcal{V}}}_0\\\mathbf{0}\\\vdots\\\mathbf{0}\\\end{matrix}\right]\in\mathbb{R}^{6n},
% {\dot{\boldsymbol{\mathcal{V}}}}_{b}=\left[\begin{matrix}{\text{Ad}}_{\boldsymbol{T}_{10}}{\dot{\boldsymbol{\mathcal{V}}}}_{0}\\\mathbf{0}\\\vdots\\\mathbf{0}\\\end{matrix}\right]\in\mathbb{R}^{6n}
% \end{equation}
% similarly, the wrench of end-effector expressed w.r.t the second-last link is, 
% \begin{equation}
% \boldsymbol{\mathcal{F}}_{t}=\left[\begin{matrix}\mathbf{0}\\\vdots\\\mathbf{0}\\{\text{Ad}}_{\boldsymbol{T}_{n+1,n}}^T\boldsymbol{\mathcal{F}}_{n+1}\\\end{matrix}\right]\in\mathbb{R}^{6n}
% \end{equation}
Given the twist of the base frame as \({\boldsymbol{\mathcal{V}}}_0\) and the wrench at the end-effector as   
\(\boldsymbol{\mathcal{F}}_{n+1}\) then then base twist expressed w.r.t link 1 and the wrench of end-effector expressed w.r.t the second last link can be achieved as
%Use adjoint representation to transform them to the closest frame in \(\{x^j_L\}_i\) and stack them up:
\begin{equation}
{\boldsymbol{\mathcal{V}}}_{b}=\left[\begin{matrix}{\text{Ad}}_{{\boldsymbol{T}}_{10}}{\boldsymbol{\mathcal{V}}}_0\\\mathbf{0}\\\vdots\\\mathbf{0}\\\end{matrix}\right]\in\mathbb{R}^{6n},
\boldsymbol{\mathcal{F}}_{t}=\left[\begin{matrix}\mathbf{0}\\\vdots\\\mathbf{0}\\{\text{Ad}}_{\boldsymbol{T}_{n+1,n}}^T\boldsymbol{\mathcal{F}}_{n+1}\\\end{matrix}\right]\in\mathbb{R}^{6n}
\end{equation}
%where \({{\boldsymbol{T}}_{10}}\) is the HTM from the base frame \(\{x^j_B\}\) to the link 1 frame \(\{x^j_L\}_1\), and \({\boldsymbol{T}_{n+1,n}}\) is the HTM from link n frame \(\{x^j_L\}_n\) to the end effector frame \(\{x^j_E\}\). For other \({{\boldsymbol{T}}_{ij}}\), it is the HTM from link j frame \(\{x^j_L\}_j\) to link i \(\{x^j_L\}_i\).

\noindent \({\boldsymbol{\mathcal{A}}}_i\) is the screw axis of joint $i$ in link frame $i$ 
%\(\{x^j_L\}_i\) 
and \({\boldsymbol{\mathcal{G}}}_1\) is the spatial inertia of link $i$ in that frame, hence they can be compactly stack together as follows:
\begin{equation}
{\boldsymbol{\mathcal{A}}}=\text{diag}({\boldsymbol{\mathcal{A}}}_1,{\boldsymbol{\mathcal{A}}}_2,\cdots,{\boldsymbol{\mathcal{A}}}_n)\in\mathbb{R}^{6n\times n}
\end{equation}
\begin{equation}
{\boldsymbol{\mathcal{G}}}=\text{diag}({\boldsymbol{\mathcal{G}}}_1,{\boldsymbol{\mathcal{G}}}_2,\cdots,{\boldsymbol{\mathcal{G}}}_n)\in\mathbb{R}^{6n\times n}
\end{equation}
The same applies to the matrices
\begin{equation}
\text{ad}_{\boldsymbol{\mathcal{V}}}=\text{diag}({\text{ad}_{\boldsymbol{\mathcal{V}}_1}},{\text{ad}_{\boldsymbol{\mathcal{V}}_2}},\cdots,{\text{ad}_{\boldsymbol{\mathcal{V}}_n}})\in\mathbb{R}^{6n\times n}
\end{equation}
% \begin{equation}
%\text{ad}_{\boldsymbol{\mathcal{A}\dot\theta}}=\left[\begin{matrix}\text{ad}_{\boldsymbol{\mathcal{A}}_1\dot\theta_1}&\mathbf{0}&\cdots&\mathbf{0}\\\mathbf{0}&\text{ad}_{\boldsymbol{\mathcal{A}}_2\dot\theta_2}&\cdots&\mathbf{0}\\\vdots&\vdots&\ddots&\vdots\\\mathbf{0}&\mathbf{0}&\cdots&\text{ad}_{\boldsymbol{\mathcal{A}}_n\dot\theta_n}\\\end{matrix}\right]\in\mathbb{R}^{6n\times 6n}
% \end{equation}
\begin{equation}
\text{ad}_{\boldsymbol{\mathcal{A}\dot\theta}}=\text{diag}(\text{ad}_{\boldsymbol{\mathcal{A}}_1\dot\theta_1},
\text{ad}_{\boldsymbol{\mathcal{A}}_2\dot\theta_2},\cdots,\text{ad}_{\boldsymbol{\mathcal{A}}_n\dot\theta_n})\in\mathbb{R}^{6n\times n}
\end{equation}
finally, having defined $\bm{\mathcal{W}}$ as
\begin{equation}\label{W}
{\boldsymbol{\mathcal{W}}}=\left[\begin{matrix}\mathbf{0}&\mathbf{0}&\cdots&\mathbf{0}&\mathbf{0}\\
\text{Ad}_{\boldsymbol{T}_{21}}&\mathbf{0}&\cdots&\mathbf{0}&\mathbf{0}\\
\mathbf{0}&\text{Ad}_{\boldsymbol{T}_{32}}&\cdots&\mathbf{0}&\mathbf{0}\\
\vdots&\vdots&\ddots&\vdots&\vdots\\
\mathbf{0}&\mathbf{0}&\cdots&\text{Ad}_{\boldsymbol{T}_{n,n-1}}&\mathbf{0}\\\end{matrix}\right]\in\mathbb{R}^{6n\times 6n}
\end{equation}
allows to write the term $\bm{\mathcal{L}}$ as follows,
\begin{equation}\label{Lterm}
{\boldsymbol{\mathcal{L}}} = (\boldsymbol{I}-{\boldsymbol{\mathcal{W}}})^{-1} = \boldsymbol{I}+{\boldsymbol{\mathcal{W}}}+{\boldsymbol{\mathcal{W}}}^2+\cdots+{\boldsymbol{\mathcal{W}}}^{n-1}
\end{equation}

\subsection{Error estimation}\label{appendix_E}
Absolute and relative positional and angular errors of the configuration are computed as follows,
\begin{equation*}
\begin{cases}
e_\theta = \arccos\left(\frac{\text{tr}(\bm{R}_e\bm{R}_t^{-1})-1}{2}\right)\\
\epsilon_\theta =\arccos\left(\frac{\text{tr}(\bm{R}_e\bm{R}_t^{-1})-1}{2}\right)\bigg/\arccos\left(\frac{\text{tr}(\bm{R}_0\bm{R}_e^{-1})-1}{2}\right)\\
e_p = ||\bm{p}_t-\bm{p}_e||\\
\epsilon_p = ||\bm{p}_t-\bm{p}_e||\bigg/||\bm{p}_e-\bm{p}_0||
\end{cases}
\end{equation*}


% \section*{ACKNOWLEDGMENT}

% The preferred spelling of the word ÒacknowledgmentÓ in America is without an ÒeÓ after the ÒgÓ. Avoid the stilted expression, ÒOne of us (R. B. G.) thanks . . .Ó  Instead, try ÒR. B. G. thanksÓ. Put sponsor acknowledgments in the unnumbered footnote on the first page.

%%%%%%%%%%%%%%%%%%%%%%%%%%%%%%%%%%%%%%%%%%%%%%%%%%%%%%%%%%%%%%%%%%%%%%%%%%%%%%%%

% References are important to the reader; therefore, each citation must be complete and correct. If at all possible, references should be commonly available publications.

\bibliographystyle{ieeetr}
\bibliography{bibbib}

\end{document}